\documentclass[10pt,letterpaper]{article}
\usepackage[top=0.85in,left=2.75in,footskip=0.75in]{geometry}

\usepackage{amsmath,amssymb}

\usepackage{changepage}

\usepackage{textcomp,marvosym}

\usepackage{cite}

\usepackage{nameref,hyperref}

\usepackage[dvipdfmx]{graphicx}

\usepackage[nopatch=eqnum]{microtype}
\DisableLigatures[f]{encoding = *, family = * }

\usepackage[table]{xcolor}

\usepackage{array}

\newcolumntype{+}{!{\vrule width 2pt}}

\newlength\savedwidth

\raggedright
\usepackage[aboveskip=1pt,labelfont=bf,labelsep=period,justification=raggedright,singlelinecheck=off]{caption}

\makeatletter
\renewcommand{\@biblabel}[1]{\quad#1.}
\makeatother

\usepackage{lastpage,fancyhdr,graphicx}
\usepackage{epstopdf}
\fancyheadoffset[L]{2.25in}
\fancyfootoffset[L]{2.25in}
\begin{document}
\vspace*{0.2in}

\begin{flushleft}
{\Large
\textbf\newline{Generating Individual Trajectories Using GPT-2 Trained from Scratch on Encoded Spatiotemporal Data} 
}
\newline
\\
Taizo Horikomi\textsuperscript{1},
Shouji Fujimoto\textsuperscript{2},
Atushi Ishikawa\textsuperscript{2},
Takayuki Mizuno\textsuperscript{3,1*}
\\
\bigskip
\textbf{1} Informatics Program, Graduate Institute for Advanced Studies, SOKENDAI, Tokyo, Japan
\\
\textbf{2} Department of Economic Informatics, Kanazawa Gakuin University, Kanazawa, Japan
\\
\textbf{3} National Institute of Informatics, Tokyo, Japan
\\
\bigskip

%
%





* mizuno@nii.ac.jp

\end{flushleft}
\section*{Abstract}
Following Mizuno, Fujimoto, and Ishikawa's research (Front. Phys. 2022), we transpose geographical coordinates expressed in latitude and longitude into distinctive location tokens that embody positions across varied spatial scales. We encapsulate an individual daily trajectory as a sequence of tokens by adding unique time interval tokens to the location tokens. Using the architecture of an autoregressive language model, GPT-2, this sequence of tokens is trained from scratch, allowing us to construct a deep learning model that sequentially generates an individual daily trajectory. Environmental factors such as meteorological conditions and individual attributes such as gender and age are symbolized by unique special tokens, and by training these tokens and trajectories on the GPT-2 architecture, we can generate trajectories that are influenced by both environmental factors and individual attributes.



\section*{Introduction}
The significance of comprehensive datasets for individual daily trajectories in decoding human mobility patterns has increased, since this approach addresses a gamut of complex issues such as natural disasters, terrorism, public safety, infectious diseases, regional disparities, promotional tactics, and traffic congestion. A detailed examination of these mobility patterns allows the identification of precipitating factors contributing to traffic bottlenecks \cite{bib1}. Moreover, we can discuss effective traffic management strategies that harmonize economic activities with infection-mitigation measures \cite{bib2,bib3,bib4}. Telecommunication providers also provide a conduit for monitoring the evacuation of people during natural calamities or mass civil unrest \cite{bib5,bib6}. The establishment of a model capable of replicating realistic attributes of individual trajectories facilitates the simulation of fluctuations in urban population movements, following the introduction of novel infrastructure, epidemic outbreak, a terrorist attack, or a global event such as the World Expo \cite{bib7,bib8,bib9,bib10}. The generation of synthetic trajectory data by this model also serves as a protective measure for individual geo-privacy \cite{bib11,bib12,bib13}.

The necessity for models that produce spatio-temporal information on an individual's daily trajectory involves the reproduction of realistic home-to-destination and destination-to-home trajectories on both routine and non-routine days \cite{bib15,bib16,bib17,bib18,bib19}. Such generative models have been developed using physics or machine learning methodologies. The former includes gravity models, preferential selection models, Markov chains, and autoregressive models \cite{bib20,bib21}, while the latter encompasses recurrent neural networks and transformer models \cite{bib14,bib22}. The machine learning methodologies provide an intricate construct with numerous parameters, thus enabling the generation of highly realistic individual trajectories. In this study, we developed a model for creating individual trajectories using the GPT-2 architecture, a transformer model that is increasingly recognized as a viable alternative to recurrent neural networks in the realm of natural language processing.

Mizuno, Fujimoto, and Ishikawa proposed a distinct location token that symbolizes a geographic location \cite{bib23} and depicted individual daily trajectories recorded at half-hour intervals as a sequence of these location tokens. This sequence was then trained from scratch using the GPT-2 architecture to create a model that generates a location every 30 minutes. However, this model has practical limitations. Most global trajectory data are expressed as origin-destination (OD) locations and the temporal intervals of their movements \cite{bib24}, and thus the temporal movement interval is not constant. To improve this point, there is need for a trajectory model capable of training and generating both locations and temporal intervals. By adding unique time interval tokens to the location tokens, we constructed a model that generates both locations and time intervals that accurately represent an individual daily trajectory.

The model in the previous study \cite{bib23} does not incorporate environmental factors such as meteorological conditions and day of the week or individual attributes such as gender and age, all of which have an impact on trajectories. We further propose a model that introduces unique tokens representing environmental and individual attributes, thereby generating trajectories that are contingent upon these environmental and individual factors.

The paper's sections are organized as follows. The ``Data'' section details the dataset on individual daily trajectories used for training the model. In the ``Related Works'' section, we present the location-generating trajectory model proposed by Mizuno, Fujimoto, and Ishikawa. The ``Models'' section expands on the model from the ``Related Works'' section, incorporating the generation of time intervals in addition to locations. The ``An Extended Model'' section further extends the model to generate trajectories that are influenced by environmental and individual attributes. Finally, the ``Conclusion'' section provides a comprehensive summary.

\section*{Data}
\subsection*{Dataset of Daily Trajectories}
The daily trajectory data exploited in this study comprises a total of 590,000 smartphones, roughly 19,000 per day, that were situated in Urayasu City, Chiba, Japan, in August 2022, courtesy of Agoop Corp. \cite{bib25}. Urayasu City is principally renowned as the location of Tokyo Disney Resort, one of the most frequented amusement parks worldwide, annually drawing over 30 million visitors. Its advantageous proximity to the metropolis, with Tokyo Station a mere 13-minute train ride away, renders it a residential area of choice. It boasts an abundance of large commercial establishments and resort hotels.

The frequency with which location data is recorded in smartphones is contingent upon the operating system and device. As of 2022, Apple's iOS enjoyed a market share of approximately 64.8\% in Japan, with Android capturing roughly 35.1\% \cite{bib26}. On iOS, the system records geographic coordinates where the user lingers for a period exceeding a certain threshold, user departure and arrival times, and, in addition, the temporal and locational information of a movement spanning a distance of approximately 500 meters to 1 kilometer. In stark contrast, Google's Android platform logs the temporal and locational information at regular time intervals. The shortest time interval documented in this dataset is 10 minutes, with the mean time interval standing at 24.27 minutes.

The Dataset of Daily Trajectories includes variables such as latitude, longitude, date and time, and unique smartphone identifiers. The precision of location information garnered via GPS is influenced by the smartphone model and the communication environment, but it typically falls within a 20-meter radius. To safeguard geo-privacy, we have omitted location data within 100 meters of each owner's inferred home coordinates (i.e., the location where the owner spends extended periods of time in the middle of the night with high frequency) and concentrated solely on the trajectories (approximately 21 million coordinates) when the owners are away from their residences. The mean duration spent outside the home is around 13 hours. Of all smartphone identifiers, approximately 71\% are linked to the owner's gender, about 64\% to the owner's age, roughly 88\% to the owner's estimated residential area (city/town level), and about 87\% to the owner's estimated work area (city/town level). To generate trajectories using the model in this investigation, the 590,000 smartphones were divided into training and testing datasets at a 4:1 ratio.

\section*{Related Works}
\subsection*{Location Generation Model by Mizuno, Fujimoto, and Ishikawa}
Mizuno, Fujimoto, and Ishikawa \cite{bib23} converted the geographical data captured in latitude and longitude into distinctive location tokens. The individual trajectories at a prescribed time resolution (e.g., every 30 minutes) are signified by a sequence of location tokens. They devised an individual trajectory model that trains these sequences of location tokens from the ground up while employing the GPT-2 architecture. This section introduces their model, which lays the foundation for the extensions in this study.

The GPT-2 architecture encompasses multiple transformer layers composed of self-attention and projection layers. This architecture operates as an autoregressive model within neural networks, sequentially predicting the subsequent token from the prior token, i.e., sequentially predicting the next location from prior locations. More specifically, the earlier work employed the GPT-2 SMALL architecture proposed by OpenAI, which consists of 12 attention heads and 12 transformer layers as well as 768 dimensions of the embedding and hidden states \cite{Radford19}. That architecture is also adopted in this paper.

We utilize the Japanese regional grid code JIS X 0410 to transform coordinates expressed in latitude and longitude into unique characters that recursively subdivide space. For instance, ``$\zeta_{1}\zeta_{2}\zeta_{3}\zeta_{4}\zeta_{5}$'' signifies the area at a 250-m resolution. Here, the character variable $\zeta_{i}$ corresponds to the $i$-th level grid code of JIS X 0410. $\zeta_{1}$ symbolizes a unique location enclosed by a square with a 40-minute difference in latitude and a 1-degree difference in longitude. $\zeta_{2}$ represents the area fashioned by dividing the first-level grid into eight equal regions in the latitude and longitude directions. Likewise, $\zeta_{3}$ denotes the area fashioned by dividing the second-level grid into ten equal regions in the latitude and longitude directions. Subsequent divisions are recursively split into two equal regions, and each $\zeta_{i}$ is assigned a unique character:

\begin{eqnarray}
\label{eq_1}
	\zeta_{i} \in \{Z_{i}|\text{Unique characters assigned to areas}\},
\end{eqnarray}

\noindent
where $Z_{i} \cap Z_{j} = \emptyset$  $(i \neq j)$. In Japan, the land areas at 250-m resolution can be represented using a unique combination of 348 (= 1st level: 176 + 2nd level: 64 + 3rd level: 100 + 4th level: 4 + 5th level: 4) characters.

The daily individual trajectory with a constant time resolution $\Delta t$ can be signified as a sequence of location information as follows,

\begin{eqnarray}
\label{eq_2}
	X(t_{0}) \_ X(t_{0} + \Delta t) \_ X(t_{0} + 2 \Delta t) \_ \cdots,
\end{eqnarray}

\noindent
where $t_{0}$ signifies the time of first departing home on a given day and the characters $X$ representing the area are $X(t)=\zeta_{1} (t) \zeta_{2} (t) \zeta_{3} (t) \zeta_{4} (t) \zeta_{5} (t)$ for 250-m resolution. The preceding and subsequent locations are connected with a ``\_'' to signify the trajectory. We append a comma character ``,'' to a temporary return home and a period character ``.'' to the final return home each day.

In a previous study \cite{bib23}, a dataset of individual trajectories recorded at $\Delta t = 30$ minute intervals in Kyoto, Japan, was transformed into the aforementioned sequence of location information, tokenized with Byte-level Byte Pair Encoding \cite{bib27} and then trained from scratch on the GPT-2 architecture. They found that the GPT-2 model can generate and predict trajectories with better precision than first-order Markov chain models, second-order Markov chain models, or CatBoost models.

\section*{Models}
\subsection*{Constructing Generative Model of Trajectory Locations and Time Intervals}
Trajectories are frequently characterized as a temporal series encompassing consecutive origin-destination coordinates and their corresponding time intervals. The time interval $\Delta t$ is non-constant. Augmenting the model described in the ``Related Works'' section, we established a model that trains and generates both locations and time intervals, and we subsequently scrutinized the generation precision of this constructed model.

\subsubsection*{Model Construction}
\begin{figure}[!h]
\begin{center}
\includegraphics[width=12cm]{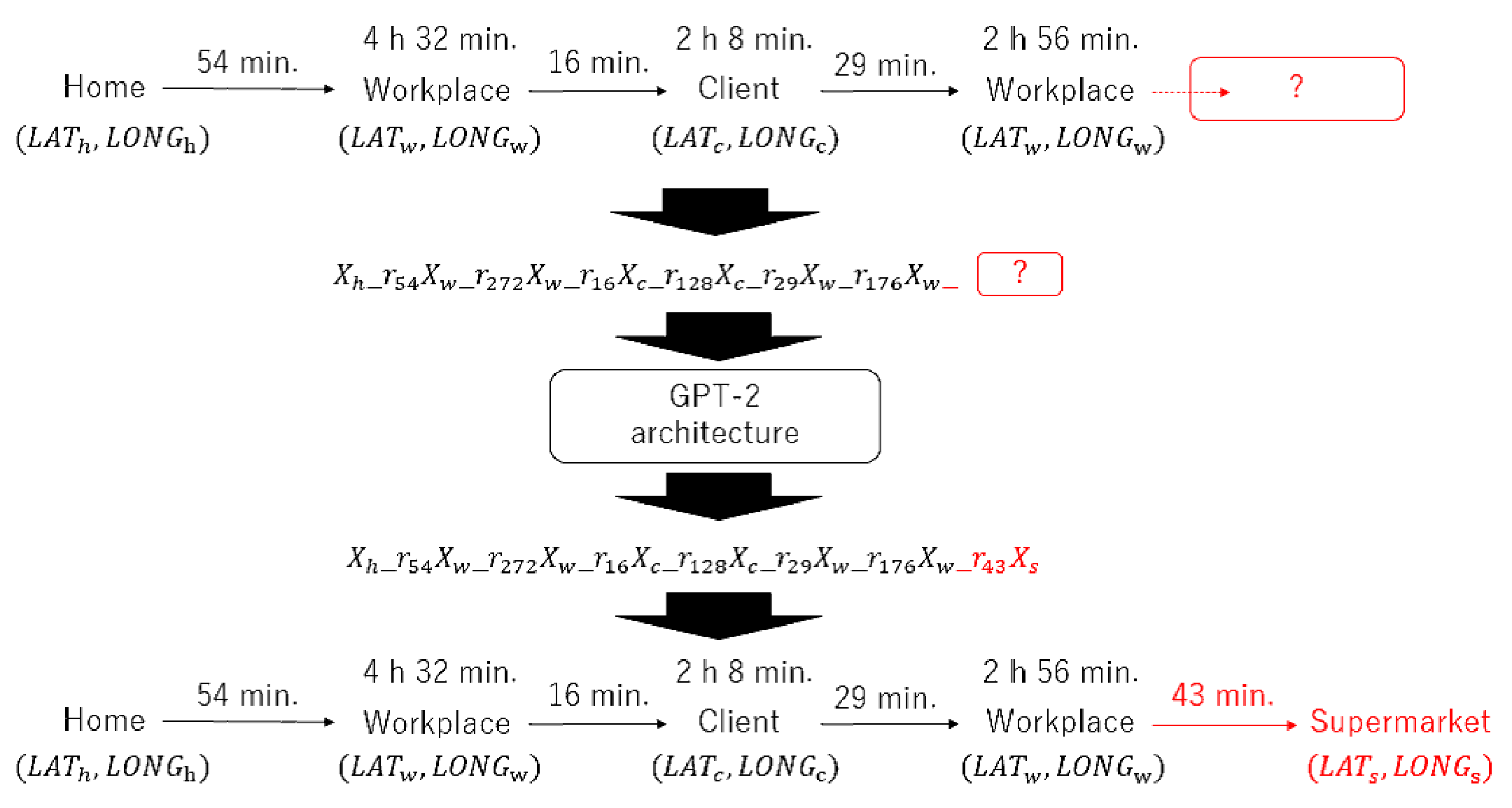}
\end{center}
\caption{{\bf Schematic representation of process used in the generation of an individual trajectory.} $LAT_{i}$ and $LONG_{i}$ express the latitude and longitude of a location $i$, respectively. Trajectories are characterized using a series of variables $X$ and $r$, which denote locations and time intervals, respectively. The schematic diagram depicts a scenario wherein the GPT-2 model, which has been trained on the characteristics of numerous trajectory sequences, anticipates that the individual will reach a supermarket 43 minutes after departure from the office.}
\label{fig_1}
\end{figure}

Figure \ref{fig_1} gives a schematic representation of the process the model uses to generate an individual trajectory. An individual travels to the workplace, undertakes tasks in the office, then proceeds to a client visit. After this business meeting, the individual returns to the office to complete internal tasks. The model anticipates the individual's subsequent destination after departure from the office. Trajectories are characterized using a series of variables $X$ and $r$, which denote locations and time intervals, respectively. The schematic diagram depicts a scenario wherein the GPT-2 model, which has been trained on the characteristics of numerous trajectory sequences, anticipates that the individual will reach a supermarket 43 minutes after departure from the office. By sequentially generating the subsequent locations and time intervals until the emergence of a ``.'' character, indicative of the end of a day, we simulate individual daily trajectories.

Here, we elucidate the definition of variable $r$, representative of a time interval. To divide 1440 minutes into several characters, we discretize the time interval $\Delta t$, with $\tau = \textrm{int} ( \log_{1.5}\Delta t ) + 1$, allocating a unique character to each ensuing discrete time interval $\tau$:

\begin{eqnarray}
\label{eq_3}
	r(\tau) \in \{R|\text{Unique characters assigned to discretized time intervals}\},
\end{eqnarray}

\noindent
where $R \cap Z_{i} = \emptyset$. By adding the character representing this discrete time interval to the characters $X$ representing the location, an individual daily trajectory can be represented by the characters $r$ and $X$ assigned to the time intervals and origin-destination coordinates, respectively, as follows:

\begin{eqnarray}
\label{eq_4}
	X(t_{0}) \_ r(\tau _{1})X(t_{0}+\Delta t_{1} ) \_ r(\tau _{2})X(t_{0} + \sum^{2}_{k=1} \Delta t_{k}) \_ \cdots,
\end{eqnarray}

\noindent
where $\Delta t_{k}$ and $\tau_{k}$ denote the time interval and the discretized time interval from the $(k-1)$-th destination to the $k$-th destination, respectively.

We implement training from scratch on the GPT-2 architecture, using the above mentioned sequence representation of the 250-m-resolution trajectories of a total of 590,000 smartphones that stayed in Urayasu City as described in the ``Data'' section. The ratio of training data to test data is 4:1. As in the ``Related Works'' section, we used the Byte Pair Encoding tokenizer and the GPT-2 architecture proposed by OpenAI, which encompasses 12 attention heads and 12 transformer layers, in addition to 768 dimensions of the embedding and hidden states.

\subsubsection*{Precision of Trajectory Generation}
\begin{figure}[!h]
\begin{center}
\includegraphics[width=12cm]{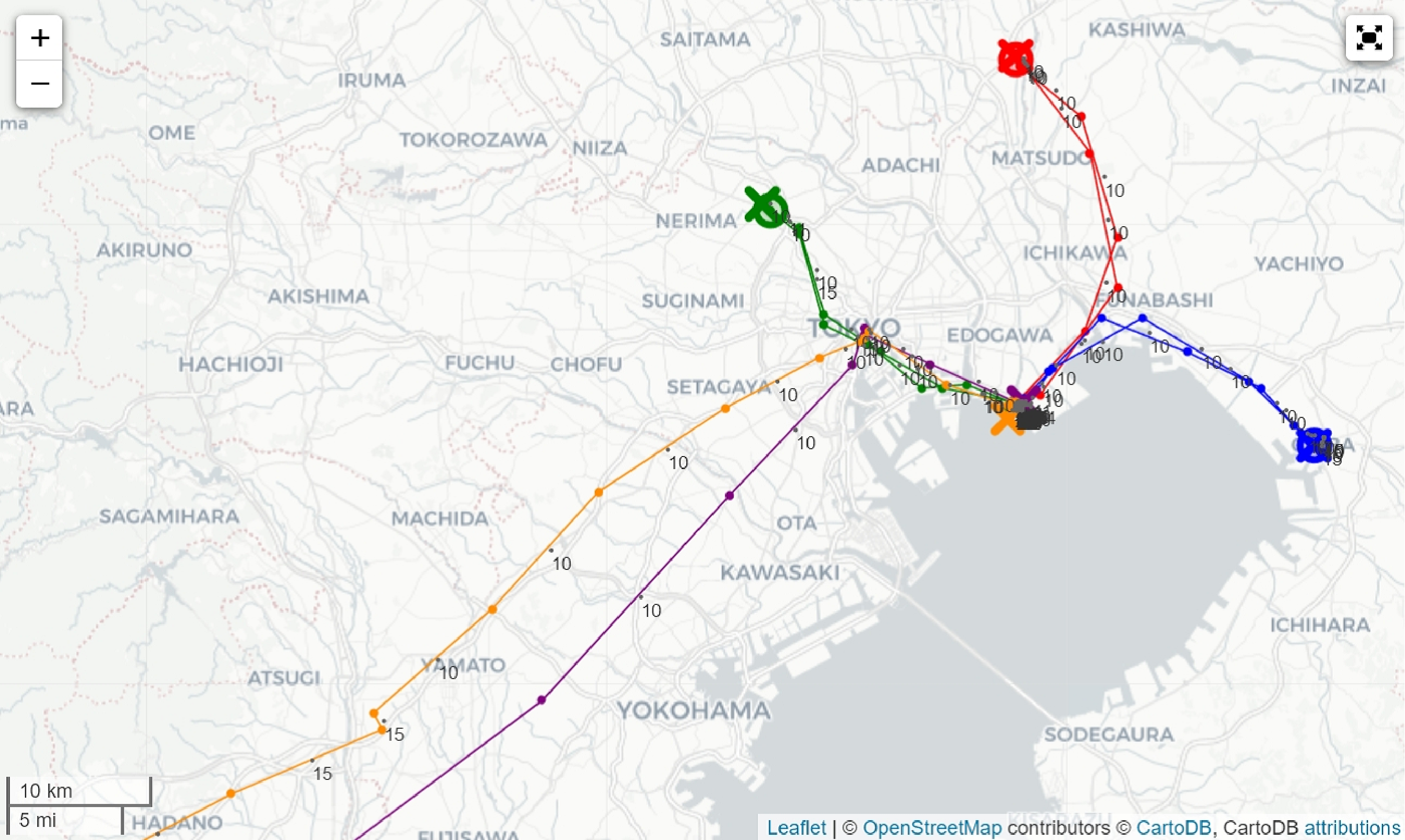}
\end{center}
\caption{{\bf Five examples of individual daily trajectories generated by the GPT-2 model in this section.} Each $\bigcirc$ represents an initial location, and each $\times$ represents a final location. The initial locations of the orange and purple trajectories were set to the locations of Nagoya City Hall and Kyoto Station, which are each more than 250 km away from Urayasu City in linear distance.}
\label{fig_2}
\end{figure}

Figure \ref{fig_2} shows five instances of individual daily trajectories generated by the trained GPT-2 model with different initial locations. With the exception of the orange and purple trajectories, indicative of long-distance travel, the final location generated by the model is close to the initial location. This typifies the daily pattern of individuals departing from their homes in the morning and returning in the evening. Such characteristics of individual trajectories have also been reported in the model proposed by Mizuno, Fujimoto, and Ishikawa. Consequently, this figure provides intuitive evidence that our model is capable of replicating trajectory characteristics in a manner comparable to their model.

Our GPT-2 model generates both locations and time intervals. We investigated the accuracy of our model with reference to the Markov chain model, which generates only locations with fixed time intervals, and the AR model, which ignores locations and generates only time intervals.

Here, we instantiate a first-order Markov chain model:

\begin{eqnarray}
\label{eq_5}
	P_{r} (X_{t}=x | X_{t - \Delta t} = x_{t - \Delta t}, \ldots, X_{1}=x_{1}, X_{0}=x_{0}) = P_{r} (X_{t}=x | X_{t - \Delta t} =x_{t - \Delta t} )
\end{eqnarray}

\noindent
and a second-order Markov chain model:

\begin{eqnarray}
\label{eq_6}
	P_{r} (X_{t}=x | X_{t - \Delta t} = x_{t - \Delta t}, \ldots, X_{1}=x_{1}, X_{0}=x_{0}) \\ \notag{ = P_{r} (X_{t}=x | X_{t - \Delta t} =x_{t - \Delta t}, X_{t - 2 \Delta t} =x_{t - 2 \Delta t} )}
\end{eqnarray}

\noindent
that generate positions at every $\Delta t$ time interval. The accuracy of the GPT-2 model in generating trajectories was compared with that of the Markov chain models. Here, $x_{t}$ represents the user's location at time $t$. In this study, the location is delineated by a grid code with a resolution of 250 m. The time interval $\Delta t$ was fixed to 30 minutes, approximately equal to the average time interval in the training data. We fitted the model to the training dataset and estimated conditional probabilities only for conditions that occur in more than 30 samples within the training dataset. For conditions that occur 30 times or fewer, we considered the conditional probabilities unknown and judged next-location generation via the Markov model to be infeasible.

\begin{figure}[!h]
\begin{center}
\includegraphics[width=12cm]{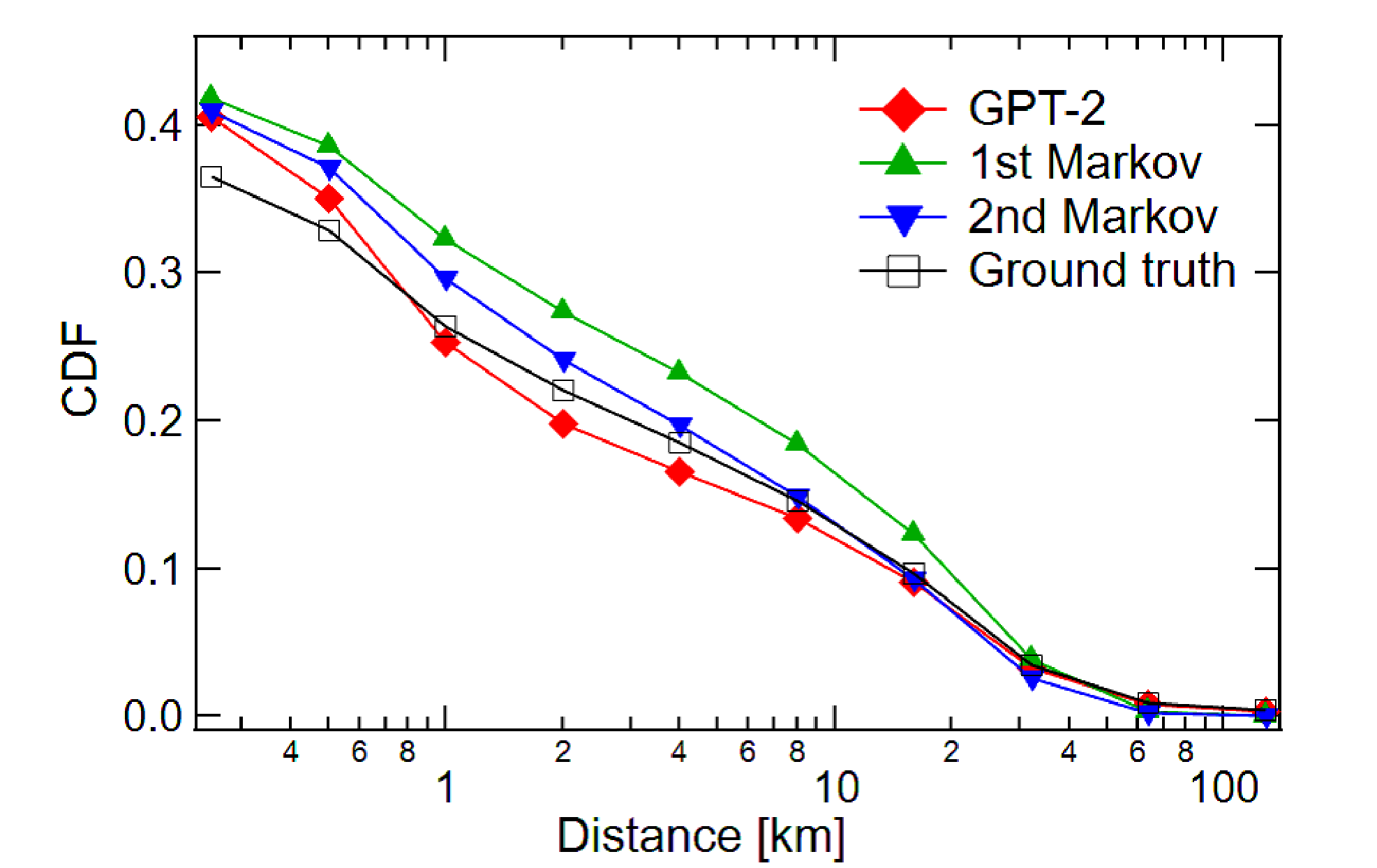}
\end{center}
\caption{{\bf Cumulative distribution on hourly moving distance in a straight line.} {\color{red}$\blacklozenge$}, {\color{green}$\blacktriangle$}, and {\color{blue}$\blacktriangledown$} are the distributions in the trajectories generated by the GPT-2 model, first-order Markov model, and second-order Markov model, respectively. $\square$ is the distribution in the actual trajectory (ground truth).}
\label{fig_3}
\end{figure}

We verified the reproducibility of trajectory characteristics using the test dataset. We input four initial locations and three initial time intervals from each trajectory in the test dataset into the GPT-2 model, and we sequentially generated the subsequent trajectory. For the first-order Markov chain model, we input one initial location, while for the second-order Markov chain model, we input two initial locations to generate the subsequent trajectory sequentially. Given that the average time spent outside of the home per day in the training data is 13 hours, we generated 26 steps (equal to 13 hours) in the Markov chain models. Figure \ref{fig_3} illustrates the cumulative distribution on the hourly moving distance in a straight line for the original and model-generated trajectories. It is evident that both the GPT-2 and Markov chain models can reproduce the distance distribution obtained from the test data. Next, we examined the probability that the prediction is within 3 km (10 km) of the actual location coordinates for the next one hour, two hours, four hours, eight hours, and the final time of the day. As demonstrated in Table \ref{table_1}, even after long periods, the GPT-2 model maintains a hit rate of nearly 20\%, whereas the hit rate of the Markov chain models significantly decays. Unlike the Markov chain models, the GPT-2 model is capable of incorporating information from locations visited many hours prior by utilizing the attention mechanism. Furthermore, the use of combinations of location tokens allows the model to handle unknown locations, corresponding to the capability of language models to accommodate unknown words. These features contribute to improving the GPT-2 model's accuracy.

Next, to compare the accuracy of the model's generation of time intervals by the GPT-2 architecture, we introduced an autoregressive model, AR($p$), with a lower bound that generates a time series of time intervals as follows: 

\begin{eqnarray}
\label{eq_7}
	\alpha_{k-1} = \phi_{0} + \phi_{1} \log(\Delta t_{k-1}) + \cdots + \phi_{p} \log(\Delta t_{k-p}) + \epsilon_{k}\\
    \notag{
      \left\{ \begin{array}{l}
      \Delta t_{k} = 10^{\alpha_{k-1}} \;\;\; ( \alpha_{k-1} \geq 1 )\\
      \Delta t_{k} = 10  \;\;\;\;\;\;\;\;\;\; ( \alpha_{k-1} < 1 )
      \end{array} \right.
     },
\end{eqnarray}

\noindent
where $\Delta t_{k}$ is the time interval from the $(k-1)$-th destination to the $k$-th destination. The lower bound was set at 10 minutes, which is the lower limit of the time interval in the training data. Using AIC, the lag order $p=3$ was estimated to be best between $p=1$ and $10$ from the training data. The coefficients $\phi$ and the probability distribution of the white noise $\epsilon_{k}$ were estimated by fitting an AR(3) model to the training data.

We then verified the ability to replicate characteristics of moving time intervals using test data. Similarly to the previous trials, we input four initial locations and three initial time intervals from each trajectory in the test data into the GPT-2 model and proceeded to sequentially generate the subsequent trajectory (locations and time intervals). For the AR(3) model, we also input three initial time intervals from each trajectory in the test data. We generated time intervals in the AR(3) model up to 13 hours (= average time spent outside of the home per day in the training data). Figure 4 presents the cumulative distribution on the time interval for the actual and model-generated trajectories. It is clear that the GPT-2 model reproduces the actual time interval distribution obtained from the test data with high precision, compared to the AR(3) model. Then, we investigated the mean absolute logarithmic errors (MALE) in the next, second, third, and fourth time intervals generated by the models compared to the actual time intervals. As shown in Table \ref{table_2}, the GPT-2 model can predict time intervals with greater precision than the AR(3) model.

\begin{figure}[!h]
\begin{center}
\includegraphics[width=12cm]{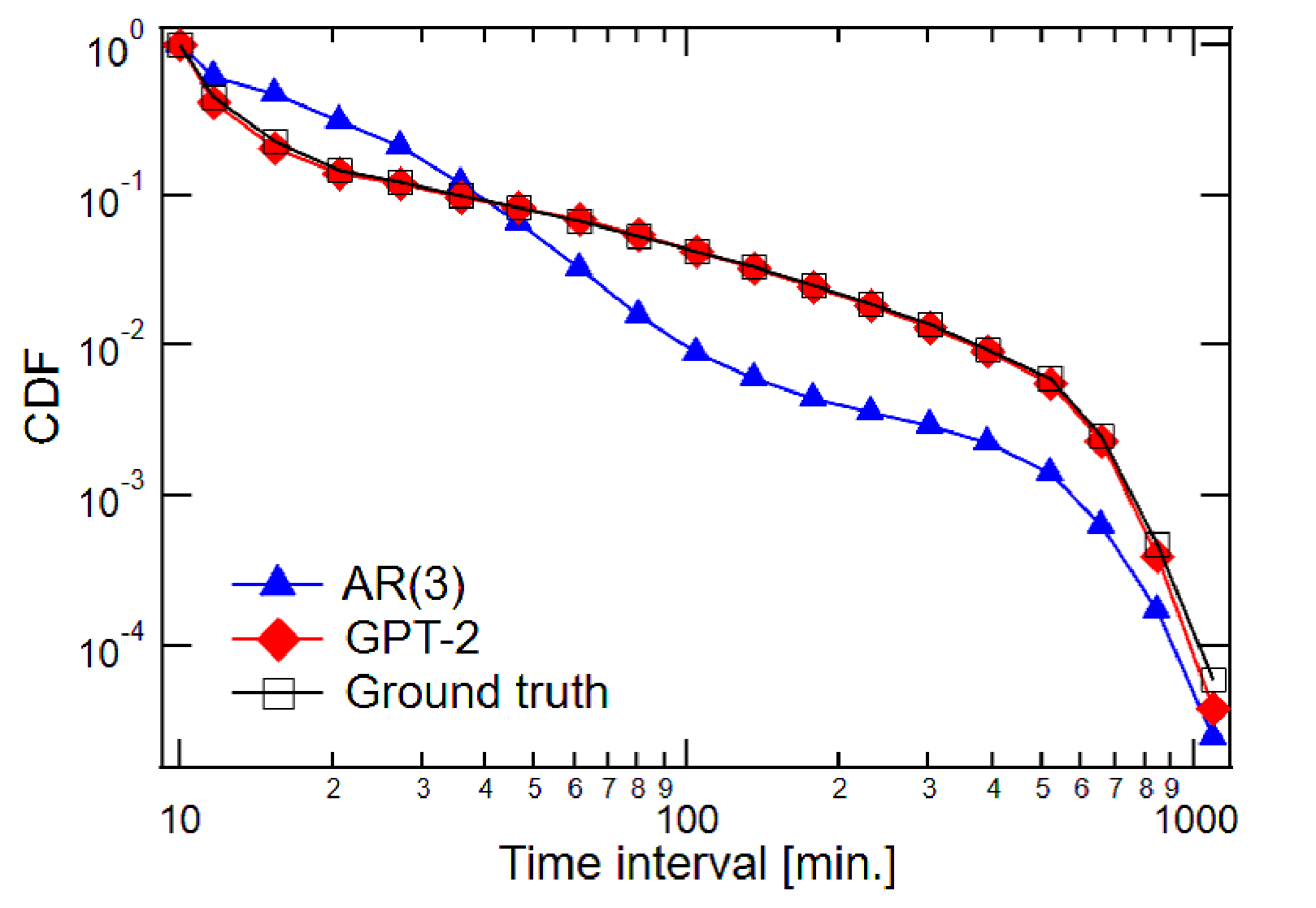}
\end{center}
\caption{{\bf Cumulative distribution on time interval.} {\color{red}$\blacklozenge$} and {\color{blue}$\blacktriangle$} are the distributions in the trajectories generated by the GPT-2 model and the AR(3) model with a lower bound, respectively. $\square$ is the distribution in the actual trajectory (ground truth).}
\label{fig_4}
\end{figure}

\begin{table}[!ht]
\begin{adjustwidth}{-2.7in}{0in} 
\centering
\caption{
{\bf Probability that the prediction is within 3 km (10 km) of the actual location coordinates for the next one hour, two hours, four hours, eight hours, and the final time of the day.} GPT-2 model with environmental factors and individual attributes is proposed in the ``An Extended Model'' section.}
\begin{tabular}{|l|l|l|l|l|l|}
\hline
Model & 1 hour & 2 hours & 4 hours & 8 hours & Final time of day \\ \hline
GPT-2 & 0.33 (0.61) & 0.28 (0.51) & 0.22 (0.42) & 0.15 (0.29) & 0.15 (0.22) \\ \hline
1st-order Markov chain & 0.20 (0.36) & 0.13 (0.28) & 0.07 (0.17) & 0.02 (0.06) & 0.00 (0.00) \\ \hline
2nd-order Markov chain & 0.25 (0.40) & 0.17 (0.30) & 0.09 (0.20) & 0.03 (0.08) & 0.00 (0.00) \\ \hline
GPT-2 with environmental factors and individual attributes & 0.36 (0.64) & 0.31 (0.54) & 0.26 (0.47) & 0.17 (0.32) & 0.15 (0.27) \\ \hline
\end{tabular}
\label{table_1}
\end{adjustwidth}
\end{table}

\begin{table}[!ht]
\begin{adjustwidth}{0in}{0in} 
\centering
\caption{
{\bf Mean absolute logarithmic errors (MALE) in the next, second, third, and fourth time intervals generated by the models compared to the actual time intervals.}}
\begin{tabular}{|l|l|l|l|l|}
\hline
Model & Next & Second & Third & Fourth \\ \hline
GPT-2 & 0.581 & 0.574 & 0.577 & 0.578 \\ \hline
AR(3) with a lower bound & 0.651 & 0.648 & 0.653 & 0.650 \\ \hline
\end{tabular}
\label{table_2}
\end{adjustwidth}
\end{table}

\section*{An Extended Model}
\subsection*{Synthesis of Individual Trajectories Influenced by Environmental Factors and Personal Characteristics}
We incorporate special tokens that embody environmental factors, such as meteorological conditions, and individual attributes such as gender and age. By concurrently training these tokens and trajectories using the GPT-2 architecture, we generate trajectories that are contingent upon both environment factors and individual attributes.

We have designated 11 special tokens across 4 categories, $s_{1}, \cdots, s_{4}$, denoting environmental factors. Special tokens signifying days of the week ([Weekday] or [Weekend]) are formulated as follows: 

\begin{eqnarray}
\label{eq_8}
	s_{1} \in \{[\text{Weekday}], [\text{Weekend}] \}.
\end{eqnarray}

\noindent
In a similar vein, the special tokens $s_{2}$, $s_{3}$, and $s_{4}$ denote temperature ([$h<25 {}^\circ$C], [$25 {}^\circ$C $\le h < 30 {}^\circ$C], [$h \ge 30 {}^\circ$C]), weather conditions ([Sunny], [Cloudy], [Rainy]), and the daily coronavirus case count ([$n<20000$], [$20000 \le n < 30000$], [$n \ge 30000$]) in Tokyo, respectively. We also prepared nine special tokens across four categories, $s_{5}, \cdots, s_{8}$, encapsulating individual attributes: gender ([Male], [Female]), age ([Under 29], [30 to 59], [Over 60]), home location ([Urayasu city(H)], [Outside of the city(H)]), and work location ([Urayasu city(W)], [Outside the city(W)]). We integrated a special delimiter token ``$|$'' to distinguish between the environmental and attribute tokens, $S=s_{1} s_{2} \cdots s_{8}$, and the initial location tokens, $X(t_{0})$. The special token sequence $S$ and the delimiter token ``$|$'' are incorporated in the individual daily trajectory, which is denoted by the sequence of time interval tokens, $r$, and location tokens, $X$, as follows:

\begin{eqnarray}
\label{eq_9}
	S|X(t_{0})\_r(i_{1})X(t_{0}+\Delta t_{1})\_r(i_{2})X(t_{0}+\sum^{2}_{k=1} \Delta t_{k})\_\cdots.
\end{eqnarray}

\noindent
As mentioned in the ``Data'' section, certain datasets lack certain individual attributes. In such instances, the corresponding special token is excluded from the sequence.

We represented all trajectories traversing Urayasu City using the aforementioned sequence and began training from scratch using the GPT-2 architecture, as implemented in previous sections. Table \ref{table_1} juxtaposes the prediction accuracy in the test data between the model proposed in the ``Models'' section, which lacks environmental factors and individual attributes, and the model presented in this section, which incorporates these factors and attributes. The probability that the prediction is within 3 km (10 km) of the actual coordinates for the subsequent one hour, two hours, four hours, and eight hours is almost always higher for the GPT-2 model with environmental factors and individual attributes than for the GPT-2 model without these factors and attributes. 

We used attention weight, denoted as $A_{i}^{(l)}$, in the $l$-th layer of token $i$ to discern salient environmental factors and individual attributes that significantly contribute to trajectory synthesis. We input environmental factors and individual attributes, four initial locations, and three initial time intervals from each trajectory in the test dataset into the GPT-2 model and sequentially generated the subsequent trajectory. In generating each destination and the corresponding time interval, we estimated the attention weights for eight special tokens representing the environmental factors, individual attributes, and aggregate attention weights for tokens representing either past locations or past time intervals. Figure 5 illustrates these attention weights for each layer in generating locations and time intervals. For both types of generation, attention to the tokens representing past locations and time intervals is noticeably higher across all layers, indicating a fundamental mechanism for anticipating future trajectories based on historical locations and time intervals. Moreover, the generations of location and travel time are not independent but rather interdependent. The attention weights for the environmental factors and individual attributes increase with the layers. Notably, in the generation of time intervals, it can be observed that in the upper layers, the daily coronavirus case count, age, weather, and gender each receive an attention weight of approximately 0.1. Information on these attributes contributes to improved prediction accuracy as shown in Table \ref{table_1}.

\begin{figure}[!h]
\begin{center}
\includegraphics[width=12cm]{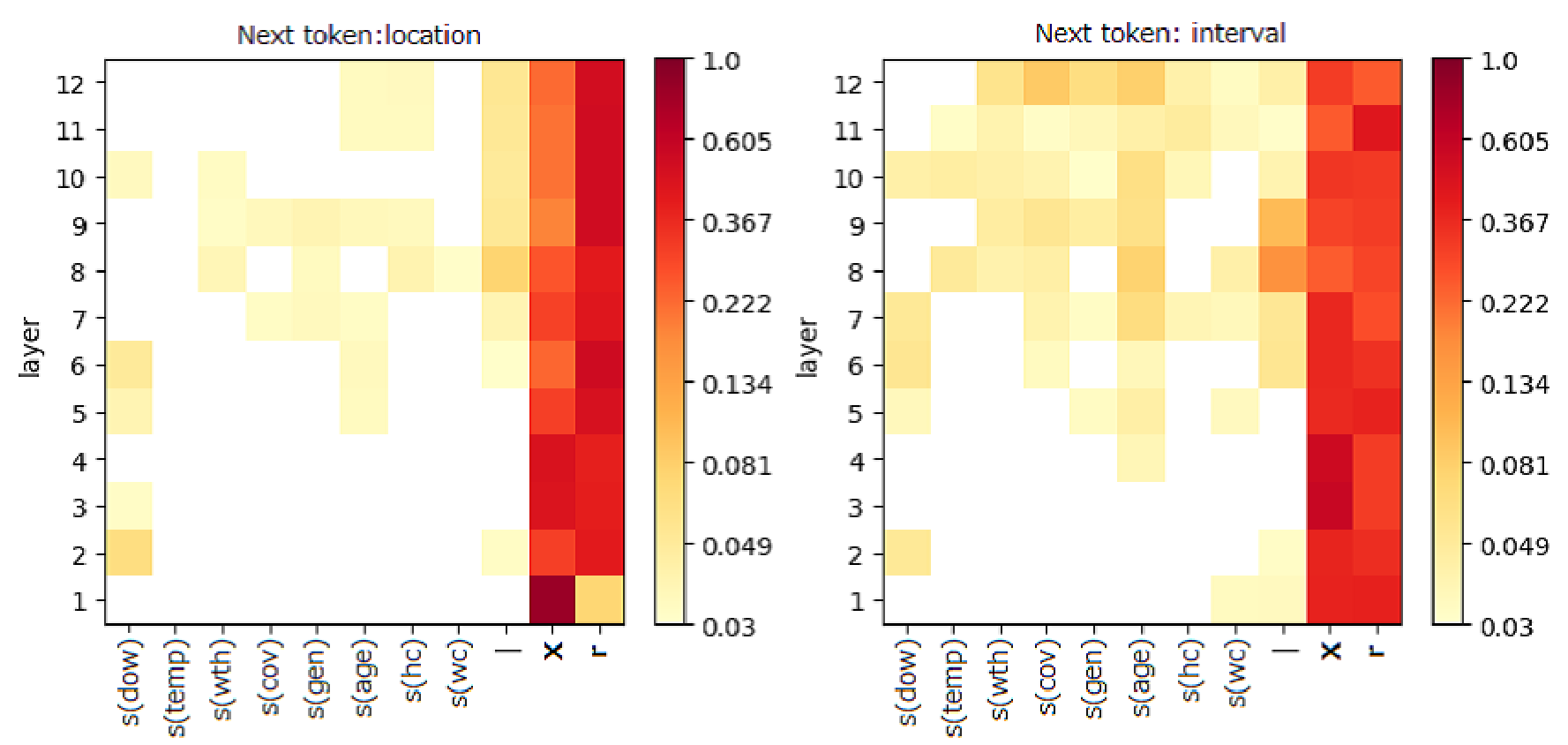}
\end{center}
\caption{{\bf Attention weights corresponding to eight special tokens, representing environmental factors, individual attributes, as well as the cumulative attention weights assigned to tokens signifying previous locations and past time intervals.} Left figure shows the attention weights for generating the next location, and right figure shows the attention weights for generating the next time interval. The vertical axis represents the attention layers. The horizontal axis, from left to right, corresponds to the following tokens: s(dow): ``day of the week,'' s(temp): ``temperature,'' s(wth): ``weather condition,'' s(cov): ``daily coronavirus case count in Tokyo,'' s(gen): ``gender,'' s(age): ``age,'' s(hc): ``home location,'' s(wc): ``work location,'' $|$: ``delimiter,'' $X$: ``past locations,'' and $r$: ``past interval'' tokens.}
\label{fig_5}
\end{figure}

\section*{Conclusion}
Emulating the approach by Mizuno, Fujimoto, and Ishikawa (Front. Phys. 2022), we transposed geographical coordinates expressed in latitude and longitude into distinctive location tokens that embody positions across varied spatial scales. In this paper, we encapsulated an individual daily trajectory as a sequence of tokens by adding unique time interval tokens to the location tokens. Using the architecture of the autoregressive language model, GPT-2, this sequence of tokens was trained from scratch, allowing us to construct a deep learning model that sequentially generates an individual daily trajectory. Environmental factors such as meteorological conditions and individual attributes, such as gender and age, are symbolized by unique special tokens, and by training these tokens and trajectories on the GPT-2 architecture, we can generate trajectories that are influenced by both environmental factors and individual attributes. To augment the predictive accuracy of the trajectory, information on the daily coronavirus case count, age, weather, and gender is efficacious.

Finally, we advocate for future research that meets the challenge of generating collective trajectories. In this study, we devised a model that omits interactions between individuals. To reproduce congestion and crowd movement, it is crucial to construct a model that incorporates the interactions. The generation of highly precise trajectories from the model can contribute to fundamental knowledge in domains such as urban planning, hypothetical scenario analysis, and computational epidemiology.



%
%
%

\end{document}